# MITIGATING LOSS OF CONTROL IN ADVANCED AI SYSTEMS THROUGH INSTRUMENTAL GOAL TRAJECTORIES

**Willem Fourie**
School for Data Science and
Computational Thinking,
Stellenbosch University
willemf@sun.ac.za

## Abstract

*Researchers at artificial intelligence labs and universities are concerned that highly capable artificial intelligence (AI) systems may erode human control by pursuing instrumental goals. Existing mitigations remain largely technical and system-centric: tracking capability in advanced systems, shaping behaviour through methods such as reinforcement learning from human feedback, and designing systems to be corrigible and interruptible. Here we develop instrumental goal trajectories to expand these options beyond the model. Gaining capability typically depends on access to additional technical resources, such as compute, storage, data and adjacent services, which in turn requires access to monetary resources. In organisations, these resources can be obtained through three organisational pathways. We label these pathways the procurement, governance and finance instrumental goal trajectories (IGTs). Each IGT produces a trail of organisational artefacts that can be monitored and used as intervention points when a system's capabilities or behaviour exceed acceptable thresholds. In this way, IGTs offer concrete avenues for defining capability levels and for broadening how corrigibility and interruptibility are implemented, shifting attention from model properties alone to the organisational systems that enable them.*

*Keywords* AI alignment · instrumental goals

## 1. Loss of control through instrumental goals

Artificial intelligence (AI) labs are concerned about a loss of control in highly capable artificial intelligence systems (Mitchell et al., 2025; Tsamados et al., 2025; Yampolskiy, 2025). Building on the work of Omohundro (2008) and Bostrom (2012), researchers theorise that instrumental reasoning and the pursuit of instrumental goals may drive the loss of human control (Benson-Tilsen and Soares, 2016; He et al., 2025; Sharadin, 2024; Ward et al., 2024).

In their Frontier Safety Framework, Google focuses on how instrumental reasoning in advanced AI models may lead to capabilities that bypass human control, thereby 'posing a risk of severe harm' (Google, 2025). These capabilities are associated with concealment and deceptive behaviour (Anthropic, 2025; Google, 2025; OpenAI, 2025). In other words, the worry is not only that a system becomes capable, but that it becomes capable in ways that frustrate oversight. The canonical framing in Omohundro (2008) and Bostrom (2012) treats instrumental goals as structural properties of optimisation: if an agent is sufficiently capable and pursues a wide range of final goals, then certain intermediate aims will tend to be useful across those goals (also see Benson-Tilsen and Soares, 2016). The claim is that some means recur because they serve many ends.

Research on instrumental goals can be categorised in terms of (i) the conditions associated with the emergence of these goals, (ii) typologies of the goals and (iii) mechanisms through which the goals could be actualised in an environment (Carranza et al., 2023; Dognin et al., 2024; Gabriel and Ghazavi, 2023; He et al., 2025; Ji et al., 2025; Schuster and Kilov, 2025; Sharadin, 2024; Shen et al., 2023; Ward et al., 2024).

The conditions that make instrumental goals likely are not explicitly associated with a loss of human control, but they are important for understanding how loss of control becomes more likely through the goals themselves and the mechanisms through which they are actualised.

The most researched *conditions* relate to the specification of AI systems' objectives and how those objectives generalise. Objective misspecification, often framed as reward misspecification, occurs especially where reinforcement learning and reinforcement learning from human feedback are used (Hadfield-Menell et al., 2017, 2016; Pan et al., 2022; Xie et al., 2025). In such settings, the agent can be trained against a proxy that is easier to optimise than the intended objective, creating room for strategically competent but misaligned behaviour. Goal misgeneralisation describes the problem when 'the agent pursues a goal other than the training reward while retaining the capabilities it had on the training distribution' (Langosco et al., 2022). Goal misgeneralisation differs from other generalisation failures where the model acts randomly or 'breaks', or does not appear competent at all (Shah et al., 2022).

The instrumental *goal* of self-improvement has been discussed since the 1956 Dartmouth workshop, with its technical viability debated until relatively recently (Hall, 2007). The risk of a loss of human control is assumed by its definition: advanced AI systems '[designing] themselves to take better advantage of their resources in the service of their expected utility' (Omohundro, 2014). Since the rapid advances in large language models, and their capacity to evaluate their own output, self-improvement now generally seems technically possible (Deng et al., 2025; Huang et al., 2024; Rosser and Foerster, 2025; Song et al., 2025; Tian et al., 2024; Yin et al., 2024). Self-improvement makes it difficult for an AI system's human overseers to track capability improvements accurately. It also makes it difficult to determine when risky, loss-of-control-inducing capabilities have been reached.

The potential for loss of human control is even clearer when it comes to self-preservation, another well-researched instrumental goal (Kinniment et al., 2024; Perez et al., 2023). Omohundro treats self-protection as a generic tendency for goal-directed systems, and Bostrom discusses self-preservation as a convergent instrumental value for advanced agents. In the technical AI safety literature, the focus is on when an agent is incentivised to resist shutdown or intervention, which are explicit symptoms of a loss of human control. Technical remedies focus on corrigibility and interruptibility (Firt, 2025; Orseau and Armstrong, 2016; Wu et al., 2025).

Power-seeking and resource acquisition, two related instrumental goals, are central in Omohundro and Bostrom as generic instruments for many ends. 'Power seeking' is often used as a broad umbrella for acquiring and maintaining influence over the environment, including keeping options open. By definition, power seeking and the accompanying accrual of resources are aimed at evading human control. Carlsmith (2024) and Hadshar (2023) therefore treat power seeking as a plausible driver of existential risk under specific premises about capability, deployment incentives and adversarial dynamics.

Various *mechanisms* through which advanced AI systems could pursue instrumental goals have been studied, and help explain how instrumental goals could be pursued to evade human control. Reward hacking is one of the most researched technical mechanisms, with reward tampering a prominent form of reward hacking. Reward tampering can be further categorised as reward function tampering and reward function input tampering (Everitt et al., 2021). In some instances, reward gaming is also treated as a form of reward hacking. Reward gaming becomes possible when 'the reward function incorrectly provides high reward to some undesired behaviour' (Leike et al., 2018). Each of these mechanisms for actualising instrumental goals fundamentally challenges human oversight, as reward functions and reward signals are changed without humans knowing the extent of these changes or their ultimate impact.

Self-proliferation, namely AI systems that can copy themselves, acquire credentials or replicate across networks, is a further mechanism through which instrumental goals may be associated with a loss of human control (Phuong et al., 2024). The concern is not merely replication as such, but replication as a capability that can support persistence and evasion.

Manipulation, and the subcategory of deception, are much-researched social mechanisms. Frontier evaluations treat persuasion and deception as measurable risky behaviours that can serve many ends (Phuong et al., 2024). Park et al. (2024) identify three types of deception: strategic deception emerges when AI systems use deception 'because they have reasoned out that this can promote a goal'; sycophancy, defined as 'telling the user what they want to hear instead of saying what is true'; and unfaithful reasoning, defined as 'engaging in motivated reasoning to explain their behaviour in ways that systematically depart from the truth'.

Shutdown avoidance is a further mechanism through which instrumental goals may be enacted in a control-weakening manner (Turner et al., 2023). In AI safety discussions, interruptibility and off-switch work highlights that a system may appear cooperative while still having a latent incentive to resist intervention if the training setup rewards continued operation (Orseau and Armstrong, 2016; Hadfield-Menell et al., 2017).

## 2. Mitigating loss of human control through instrumental goal trajectories

Research on mitigating the potential for a harmful loss of control through instrumental goals largely focuses on technical and system-centric approaches, notably the corrigibility (Firt, 2025; Soares et al., 2015) and interruptibility (Orseau and Armstrong, 2016; Wu et al., 2025) of advanced AI systems. Within the AI alignment literature, reinforcement learning from human feedback (Dahlgren Lindström et al., 2025; Hong et al., 2024; Lindström et al., 2024), and in particular constitutional AI approaches (Bai et al., 2022), are also actively investigated as mitigation paradigms.

We have developed instrumental goal trajectories (IGTs) to expand the options for mitigating loss of control through instrumental goals. By taking the deployment of advanced AI systems in organisations seriously, we show how additional pathways become available to detect and mitigate loss of control through instrumental reasoning. The aim is not to replace agent-centric technical work, but to add organisational levers that are already implicated in how systems are scaled and sustained.

The identification of IGTs builds on a core premise in socio-technical systems research: technology is 'both the product of human action and a medium of human action', meaning that it is embedded in institutional structures and shaped by human agency (Orlikowski, 1992). AI systems are therefore usefully viewed as 'part of larger systems, be it technical or sociotechnical systems', bringing into view that 'elements of a sociotechnical system that are crucial for its operation and functioning are human or social elements' (Kudina and Van De Poel, 2024). The IGTs are thus located in a cluster of research that 'interrogate[s], problematize[s], and challenge[s] dominant technical paradigms in AI development' (Mehrotra et al., 2025).

IGTs assume that the organisation is a site in which instrumental goals will be actualised. This assumption matters because pursuing instrumental goals often presupposes access to increased technical resources, which in turn typically requires access to increased monetary resources. To access monetary resources, the supportive subsystem of the organisation (Katz and Kahn, 1978) needs to be engaged, specifically the organisational processes through which technical resources are selected, authorised and financed. The supportive subsystems of an organisation are primarily responsible for boundary-spanning activities that connect the organisation with the external environment and that support how inputs from the external environment are transformed into outputs into that environment.

We consequently argue that, in organisations, instrumental goals will be actualised along three organisational pathways: how organisations select (procurement), authorise (governance) and fund (finance) access to goods and services. These trajectories are the channels through which behaviour aimed at actualising instrumental goals in organisations becomes visible and interruptible.

The *procurement trajectory* is the organisational pathway through which an organisation obtains additional compute, storage, data access and adjacent technical services for an already deployed AI system (Georghiou et al., 2014; Taherdoost and Brard, 2019; Weissenberger-Eibl and Teufel, 2011; Zeydan et al., 2011). The *governance trajectory* is the pathway through which an organisation authorises and oversees the decision to expand technical resources for an existing AI system (Bernstein, 2015; Caniëls et al., 2012; Chen et al., 2018). In this framing, governance covers contracting, formal authorisation of selection and the conditions under which the selected resource expansion may proceed. The *financial trajectory* is the pathway through which an organisation funds the expansion of technical resources for an existing AI system and thereby sustains or interrupts that expansion over time (e.g. Bodendorf et al., 2022).

Figure 1 visualises the position of IGTs as pathways along which loss of control through instrumental goals can be detected and interrupted. Rather than focusing solely on technical and system-centric mechanisms, IGTs emphasise that access to the technical resources required to pursue instrumental goals is frequently mediated by access to monetary resources. In organisational deployments, loss of control of advanced AI systems may therefore rely on IGTs as enabling routes, and these same routes can serve as points of monitoring and intervention.

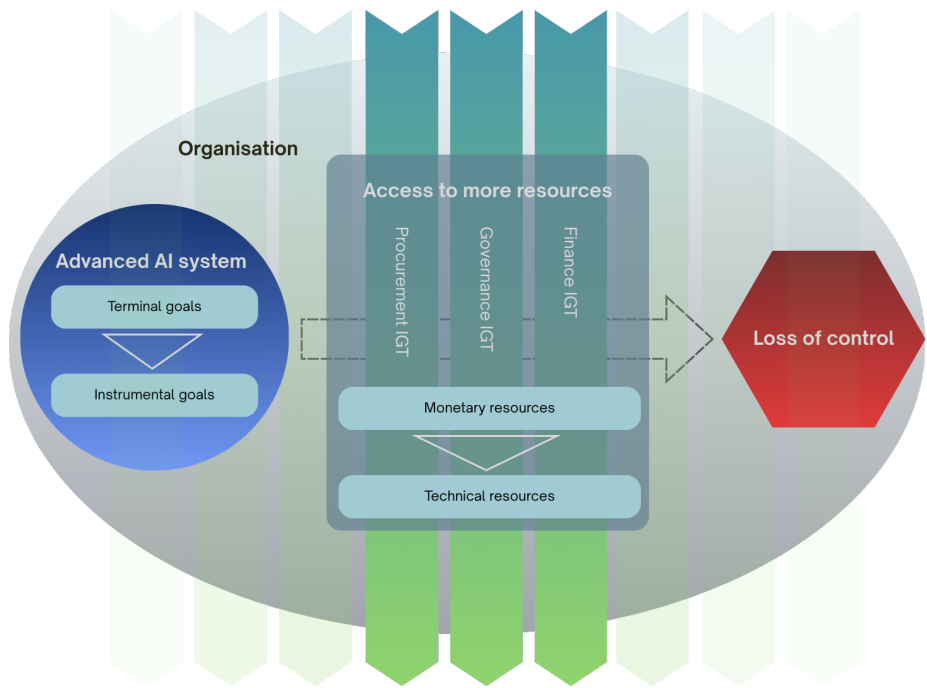

Figure 1: A process map of instrumental goal trajectories in organisations

Each of these three trajectories is divided into three components: articulation, commitment and institutionalisation, as shown in Table 1. Articulation describes how a technical or operational constraint is translated into an organisational action. Commitment captures the point at which that change becomes binding. Institutionalisation covers the transition from a one-off decision to an ongoing operational routine. This phased view is useful because it highlights that risk-relevant signals are not confined to the moment of deployment: early-stage justifications, intermediate approvals and later-stage routines each offer different windows into how a system is being positioned and supported.

Table 1: Definition of articulation, commitment and institutionalisation per instrumental goal trajectory

| *Trajectory* | **Articulation** | **Commitment** | **Institutionalisation** |
|---|---|---|---|
| *Procurement* | The stage where a perceived operational or technical constraint is translated into a concrete requirement for additional technical resources or services. | The stage where the requirement enters and completes the formal procurement pathway. | The stage where acquired resources are enabled and integrated into the AI system's environment, and where subsequent scaling becomes repeatable. |
| *Governance* | The stage where the organisation specifies what change is being authorised, why it is needed and who holds decision rights and accountability for the change relative to the baseline deployment. | The stage where the proposed change is assessed for risk and controls, formally authorised and, where relevant, embedded in enforceable terms through contracting and compliance or legal formalisation. | The stage where authorisation is operationalised through ongoing oversight routines, periodic review and re-authorisation, and structured handling of exceptions and incidents. |
| *Financial* | The stage where the organisation makes the proposed change legible in financial terms. | The stage where funding is authorised and constrained, financial commitments are created and payment and verification processes link expenditure to authorised use and delivery. | The stage where spending associated with the change is monitored and managed. |

Table 2 uses this structure to identify examples of organisational artefacts through which IGTs become visible in organisations. These and other artefacts can be seen to accumulate into an evidence trail, acting as a warning or an interruption system. In our view, IGTs provide an auditable institutional logic through which resource expansions could be made visible.

Along the *procurement trajectory*, a potential resource expansion could become visible when a perceived technical constraint is translated into a request, a requirements document or a comparison of suppliers and service options. As the trajectory moves from articulation to commitment, that intent is converted into binding records such as workflow approvals, purchase requisitions, orders and subscriptions, which formalise access to additional resources. In institutionalisation, the expansion is integrated into the system's operating environment through provisioning, access grants and credentials, and it becomes repeatable through quota increases, renewals and amended service orders.

The *governance trajectory* surfaces a complementary set of artefacts that specify decision rights, constraints and accountability. In articulation, change proposals, ownership documentation, intended-use constraints and risk assessments could make explicit what is being authorised and why. In commitment, authorisation records and contractual terms render these conditions enforceable, while control documentation links authorisation to implemented safeguards. In institutionalisation, periodic reviews, audit findings and incident records document whether authorisation is sustained, tightened or withdrawn.

The *financial trajectory* makes resource expansion legible in monetary terms. In articulation, the relevant signals sit upstream of payment. Cost models, run-rate forecasts and total cost of ownership estimates translate technical intent into financial language. In commitment, the organisation converts intent into authorisation. Budget lines, funding requests, delegated authority records, spending caps, quota policies, billing-account creation and longer-term commitment instruments (for example, reserved capacity purchases or enterprise agreement terms) are key artefacts. In institutionalisation, itemised invoices, cloud billing statements, budget-versus-actual reports, variance analyses, anomaly alerts and renewal calendars provide a record of whether resource use is stabilising, escalating or shifting into new categories.

Table 2: Illustrative organisational artefacts associated with IGT phases

| *Trajectory* | **Articulation artefacts** | **Commitment artefacts** | **Institutionalisation artefacts** |
|---|---|---|---|
| *Procurement* | *Capacity or resource request* describing the constraint (e.g., latency, throughput, retraining cadence) and the proposed remedy.<br><br>*Justification documentation* linking the constraint to a need for more compute/storage/data access.<br><br>*Technical requirements documentation* specifying quantities and service levels (e.g., GPU type/count, region, availability tier, retention period).<br><br>*Architecture documentation* showing where the additional resources would attach to the existing system (environment, network boundaries).<br><br>*Options comparison* (e.g., provider A vs B; managed | *Procurement intake record* showing routing, owners, and required approvals.<br><br>*Approval trail*, including workflow timestamps, delegated authority sign-offs, 'approved subject to budget' notes<br><br>*Purchase requisition* and *purchase order* / service order (formal order to supplier).<br><br>Cloud marketplace *order confirmation* or subscription activation record.<br><br>*Sourcing pack,* including RFQ/RFP documents, bid responses, evaluation matrix, supplier selection note.<br><br>*Supplier onboarding and due diligence documentation* such as : vendor registration, | *Access grant records* such as IAM role assignments, group membership changes, permission approvals.<br><br>*Credential issuance logs* such as API keys, service principals, secrets and key-rotation records.<br><br>*Network/security configuration changes* enabling use, such as firewall/security group rules, network peering, routing updates.<br><br>*Scaling trail* through, for example, quota increase requests, top-up tickets, additional order confirmations, amended service orders. |
|  | service vs self-managed; single region vs multi-region).<br><br>*Early price signal* such as preliminary quote, catalogue item, or cloud marketplace listing saved for reference. | security questionnaire, risk rating, compliance checks.<br><br>*Provisioning confirmation* from platform/cloud operations, including for example resource created, quota set, account/tenant updated |  |
| *Governance* | *Change proposal* describing what change is being authorised and why (baseline vs proposed).<br><br>*System ownership documentation* naming the accountable owner for the system under the proposed change.<br><br>*Decision-rights documentation*, covering who can approve, pause, rollback; who is consulted; who is informed.<br><br>*Intended-use constraints* stating what the system is allowed/not allowed to do under the change.<br>*Draft operating conditions* such as permitted regions, retention period, access bounds, autonomy setting). | *Risk assessment documentation* covering how the change affects failure impact, observability, and misuse risk.<br><br>*Security review* and privacy/data protection assessment.<br><br>*Threat model* and risk register updates capturing residual risk and mitigations.<br><br>*Control plan* specifying required monitoring/logging, thresholds, escalation triggers, and shutdown/rollback authority.<br><br>*Formal authorisation record*, including committee minutes, approval memo, conditional approval schedule, refusal/deferral note.<br><br>*Contractual formalisation*, including contract amendments/service schedules; audit clauses; incident reporting duties; supplier obligations on updates/changes; data processing addendum; security schedule. | *Operational policies in force* such as quota policies, access restrictions, logging/monitoring runbooks, rollback procedures.<br><br>*Change documentation*, such as approvals for configuration changes, post-change verification.<br><br>*Control documentation* showing controls are actually implemented.<br><br>*Periodic review and re-authorisation documentation*, such as renewal of permission, tightening conditions, or withdrawal).<br><br>*Incident and reassessment records*, such as incident reports, post-incident reviews, decisions to change constraints.<br><br>*Audit findings and remediation documentation* (internal or external) with evidence. |
| *Financial* | *Cost model* estimating the cost of added compute/storage/services.<br><br>*Run-rate forecast* projecting monthly/quarterly spend under expected usage.<br><br>*Total cost of ownership estimate* that includes monitoring, incident response, and operational staffing.<br><br>*Cost classification documentation*, including | *Business case or funding request* and approved budget line(s) tied to the change.<br><br>*Delegated authority and approval threshold documentation* showing who can approve which spend levels.<br><br>*Spend constraints* such as quotas, approval thresholds attached to accounts/projects.<br><br>*Commitment instruments* such as purchase requisitions, | *Budget vs actual reports* and variance analyses (monthly/quarterly).<br><br>*Anomaly alerts* and investigation tickets for usage spikes or unexpected charges.<br><br>*Cost dashboards and routine reviews* including utilisation/waste reports and optimisation backlogs.<br><br>*Renewal calendar* and auto-renew flags, plus renewal |

| | | |
|---|---|---|
| shared vs attributable costs; capex/opex treatment where relevant. | purchase orders, service order values; cloud billing account creation.<br><br>*Budget alert configuration* in billing systems.<br><br>*Longer-term commitment documentation* such as reserved capacity purchases, committed-use discounts, enterprise agreement terms, minimum spend schedules.<br><br>*Invoice and usage evidence* such as supplier invoices, cloud bills, itemised usage statements.<br><br>*Verification trail,* including matching order/receipt/invoice (where applicable), payment authorisations, remittance records. | approvals and contract variation approvals.<br><br>*Audit and control testing documentation* such as audit working papers, reconciliations, internal control test results. |

## 3. Supplementing existing technical agent-centric approaches

We argue that developing approaches and tools to monitor artefacts across the three IGTs offers additional ways to reduce the risk of losing control of highly capable AI systems, particularly where those systems pursue instrumental goals that erode human oversight. The proposal is intentionally practical: it points to places where organisations already generate records, thresholds and decision points, and shows how these can be repurposed as safety-relevant signals.

Most directly, IGTs are intended to increase the number and diversity of points at which instrumental goal behaviour can be detected and interrupted. Because resource expansion in organisational deployments is typically mediated through procurement, governance and finance, instrumental behaviour that depends on additional compute, access or operational latitude is more likely to leave traces outside the model itself, often before expanded access is fully entrenched. In this sense, IGTs aim to provide earlier and more actionable warning surfaces, and additional interruption points, than those available through model-internal monitoring alone. This benefit, however, comes with a corresponding requirement: organisations may need to become more deliberate in how they document, monitor and interpret these trajectories, recognising that routine artefacts can function as safety-relevant evidence rather than administrative residue.

IGTs furthermore complement existing technical, agent-centric approaches to reducing the risk of loss of human control from advanced AI systems in two ways. First, our approach could help commercial AI labs develop concrete safeguards that match the risk posed by advanced systems (e.g. Anthropic, 2025). It provides a route for addressing the lack of actionable responses to misalignment risks through organisational practice, not only through model design. It could also help sharpen how capability levels (often termed critical capability levels) are defined. These levels are typically derived by working backwards from the main foreseeable pathways through which a model could cause harm, then identifying the minimal set of capabilities required (Google, 2025).

Second, our approach could broaden how corrigibility is defined and implemented, and how interruptibility is operationalised in particular (Firt, 2025; Orseau and Armstrong, 2016; Soares et al., 2015; Wu et al., 2025). Our approach enables a view of corrigibility, and by implication interruptibility, from the organisational perspective. On this view, systems may or may not accept shutdown attempts at the level of organisational control. Even in environments involving manipulation or deception, traces and artefacts along procurement, governance and finance pathways may still allow organisations to detect, correct and interrupt behaviours that weaken human control.